\documentclass[lettersize,journal]{IEEEtran}
\usepackage{amsmath,amsfonts}
\usepackage{algorithmic}
\usepackage{algorithm}
\usepackage{array}
\usepackage[caption=false,font=normalsize,labelfont=sf,textfont=sf]{subfig}
\usepackage{textcomp}
\usepackage{stfloats}
\usepackage{url}
\usepackage{verbatim}
\usepackage{graphicx}
\usepackage{cite}
\usepackage{times}
\usepackage{epsfig}
\usepackage{graphicx}
\usepackage{amsmath}
\usepackage{amssymb}

\usepackage{booktabs}
\usepackage{multirow}
\usepackage{amssymb}
\usepackage{amsmath}
\usepackage{multirow}
\usepackage{newfloat}
\usepackage{listings}
\usepackage{color}
\usepackage{tablefootnote}

\begin{document}

\title{Revisiting and Exploring Efficient Fast Adversarial Training via LAW: Lipschitz Regularization and Auto Weight Averaging}

\author{Xiaojun Jia,~\IEEEmembership{Student Member,~IEEE,}
        Yuefeng Chen,
        Xiaofeng Mao,
        Ranjie Duan,
        Jindong Gu,
        Rong Zhang, \\
        Hui Xue
        and~Xiaochun Cao~\IEEEmembership{Senior Member,~IEEE}
\IEEEcompsocitemizethanks{

\IEEEcompsocthanksitem Xiaojun Jia is with
State Key Laboratory of Information Security,
Institute of Information Engineering, Chinese Academy of Sciences, Beijing 100093, China, and also with School of Cyber Security, University of Chinese Academy of Sciences, Beijing 100049, China.
(e-mail: jiaxiaojun@iie.ac.cn)
\IEEEcompsocthanksitem Yuefeng Chen, Xiaofeng Mao, Ranjie Duan, Rong Zhang and Hui Xue are with the Security Department of Alibaba Group. (e-mail: yuefeng.chenyf@alibaba-inc.com, mxf164419@alibaba-inc.com, ranjieduan@gmail.com, stone.zhangr@alibaba-inc.com, hui.xueh@alibaba-inc.com)
\IEEEcompsocthanksitem Jindong Gu is with Department of Engineering Science in the University of Oxford, Oxford OX1 3PJ, United Kingdom. (e-mail:  jindong.gu@outlook.com)
\IEEEcompsocthanksitem Xiaochun Cao (Corresponding) is with School of Cyber Science and Technology, Shenzhen Campus, Sun Yat-sen University, Shenzhen 518107, China (e-mail: caoxiaochun@mail.sysu.edu.cn)
\IEEEcompsocthanksitem  Xiaochun Cao is the corresponding author
}
}

\markboth{Manuscript for IEEE Transactions on Image Processing}%
{Shell \MakeLowercase{\textit{et al.}}: Bare Demo of IEEEtran.cls for IEEE Journals}


\maketitle

\begin{abstract}

Fast Adversarial Training (FAT) not only improves the model robustness but also reduces the training cost of standard adversarial training. However, fast adversarial training often suffers from Catastrophic Overfitting (CO), which results in poor robustness performance. 
 Catastrophic Overfitting describes the phenomenon of a sudden and significant decrease in robust accuracy during the training of fast adversarial training.
Many effective techniques have been developed to prevent Catastrophic Overfitting and improve the model robustness from different perspectives. However, these techniques adopt inconsistent training settings and require different training costs, i.e, training time and memory costs, leading to unfair comparisons. In this paper, we conduct a comprehensive study of over 10 fast adversarial training methods in terms of adversarial robustness and training costs. We revisit the effectiveness and efficiency of fast adversarial training techniques in preventing Catastrophic Overfitting from the perspective of model local nonlinearity and propose an effective Lipschitz regularization method for fast adversarial training. Furthermore, we explore the effect of data augmentation and weight averaging in fast adversarial training and propose a simple yet effective auto weight averaging method to improve robustness further. By assembling these techniques, we propose a \textbf{FGSM}-based fast adversarial training method equipped with \textbf{L}ipschitz regularization and \textbf{A}uto \textbf{W}eight averaging, abbreviated as FGSM-LAW. Experimental evaluations on four benchmark databases demonstrate the superiority of the proposed method over state-of-the-art fast adversarial training methods and the advanced standard adversarial training methods.

\end{abstract}


\section{Introduction}
\label{sec:summarize}
Deep Neural Networks (DNNs)~\cite{deng2013new,sainath2013improvements,krizhevsky2017imagenet,zhao2019object} can be easily fooled by Adversarial Examples (AEs)~\cite{dong2018boosting,fan2020sparse,jia2020adv,duan2020adversarial,wang2021enhancing}which are carefully crafted by adding imperceptible perturbations to benign samples. 
Adversarial Training (AT) has been demonstrated as one of the most effective methods to improve the model robustness against adversarial examples.
In Standard Adversarial Training (SAT) \cite{cai2018curriculum,wang2019improving,wang2021convergence,zhang2020attacks,cui2021learnable,sitawarin2021sat,mao2022towards}, multi-step attack methods such as Projected Gradient Descent (PGD) are often applied to generate adversarial examples, which are computationally costly. This makes it challenging to apply AT on large-scale datasets.
To reduce training expenses, Fast Adversarial Training (FAT) \cite{tramer2017ensemble} has been proposed,which uses a single-step attack such as Fast Gradient Sign Method (FGSM) to approximate the multi-step attack for adversarial example generation. 
Recent work \cite{wong2020fast} has revealed that FGSM-AT does not increase the model's resilience against adversarial examples. In detail, Wong \emph{et al.}\cite{wong2020fast} first discover that FGSM-based AT is prone to Catastrophic Overfitting (CO), 
where model robustness initially increases during FGSM-AT but then sharply decreases after a few training epochs.

Some fast adversarial training variants  \cite{kim2021understanding,jia2022boosting,li2022subspace,de2022make} have been proposed to prevent Catastrophic Overfitting and improve the model robustness.
 Sample initialization ~\cite{wong2020fast,jia2022boosting}  and loss regularization ~\cite{sriramanan2020guided} are two commonly used strategies to prevent Catastrophic Overfitting and improve the model robustness. 
Sample initialization ~\cite{zheng2020efficient,jia2022prior} provide better initialized perturbations of FGSM for improving the quality of the adversarial examples. 
Loss regularization~\cite{sriramanan2021towards} adopts regularity to guide the generation of stronger adversaries and smooths the loss surface for better robustness. 
Although these techniques can alleviate Catastrophic Overfitting effectively, 
some of them may introduce more training burden,
violating the original intention of fast adversarial training. 
These inconsistent training time requirements can also lead to unfair comparisons.
\begin{figure*}[t]
\begin{center}
   \includegraphics[width=0.9\linewidth]{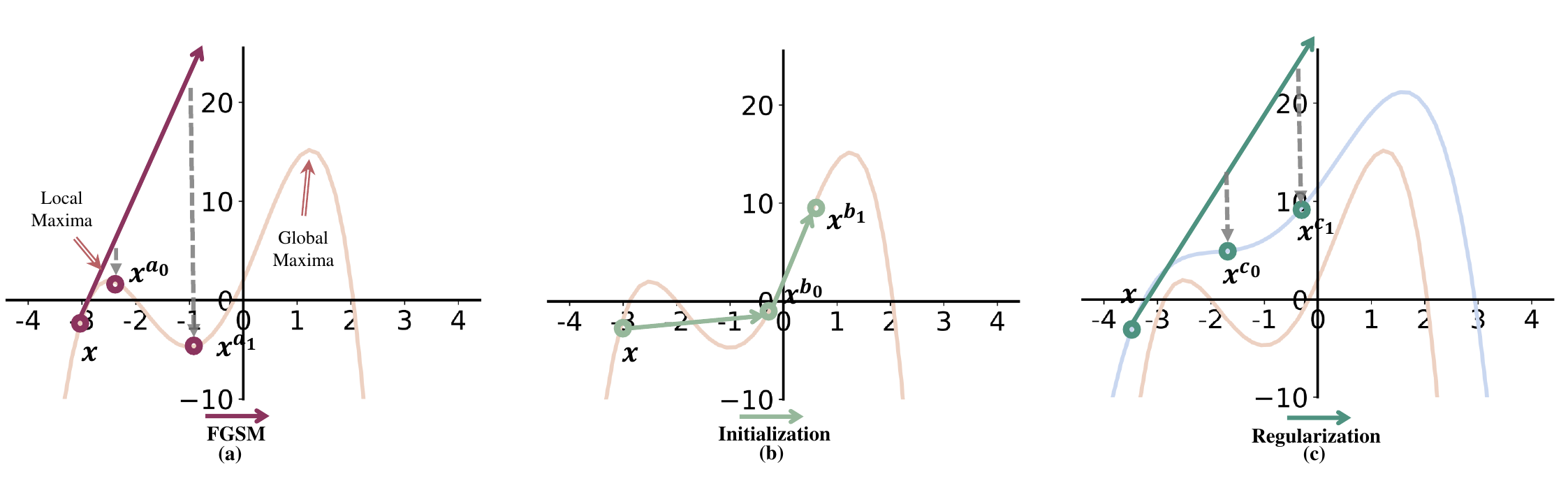}
\end{center}
\caption{ 
Generation process of adversarial examples generated by FGSM in FAT on the maximization loss function. (a) Using a zero initialization. (b) Using the sample initialization. (c) Using the regularization.
}
\label{fig:CO}
\end{figure*}

\par To conduct a comprehensive study, 
we investigate the used techniques of over 10 papers on the fast adversarial training methods, which are summarized in Supplementary Material.
We find the robustness improvement of these techniques is still correlated with training costs even in fast adversarial training.
For example, Andriushchenko \emph{et al.}~\cite{andriushchenko2020understanding} propose a gradient alignment regularization method that can relieve Catastrophic Overfitting and enhance model robustness. But calculating the regularization term requires a significant amount of time. 
 And Zheng \emph{et al.}~\cite{zheng2020efficient} propose to use the adversarial perturbations from the last epoch as the sample initialization to improve  model robustness, 
 but this method requires extra computational costs and high memory consumption to restore previous perturbations.
 Therefore, conducting empirical studies to evaluate different FAT-related techniques quantitatively would be essential.

\par In this paper, we focus on exploring and finding efficient and effective FAT-related strategies. We roughly divide existing FAT-related techniques into two categories: sample initialization and regularization. In addition, we also explore the effect of several techniques on the robustness improvement for fast adversarial training, \emph{i.e,} data augmentation, and weight averaging. 
We conduct extensive experiments to evaluate the effectiveness and computational cost of various techniques in each category and conclude the following findings:
\begin{itemize}
    \item \textbf{Sample Initialization} Prior/learning-based initialization prevents Catastrophic Overfitting and achieves better model robustness, which requires more training costs. Random sample initialization with an appropriate step size also prevents Catastrophic Overfitting but achieves limited model robustness improvement without extra training cost.
    
     \item \textbf{Regularization} Regularization is unnecessary during the inner maximization step in fast adversarial training, but regularization plays a critical role in the outer loss minimization step in fast adversarial training.

    \item \textbf{Data Augmentation} Different from SAT~\cite{rebuffi2021fixing}, fast adversarial training can achieve better model robustness  when equipped with several sophisticated data augmentations.
    
     \item \textbf{Weight Averaging} In contrast to SAT~\cite{wang2022self}, original fast adversarial training (FGSM-RS) cannot boost robustness through the mere implementation of weight averaging technique.

\end{itemize}

Based on the findings above, 
\textbf{a)} for sample initialization, considering the time consumption, we adopt random initialization. 
\textbf{b)} For regularization, motivated by Lipschitz constraint, we propose a novel regularization approach only use in minimization optimization to achieve better performance and less time overhead.
\textbf{c)} For data augmentation, we find data augmentation can improve the quality of adversarial examples and adopt the Cutout for fast adversarial training. 
\textbf{d)} For weight averaging, we discover that the reason WA fails on original fast adversarial training (FGSM-RS) is that model parameters accumulate non-robust parameters trained on the low-quality adversarial examples. Then we also  propose a simple yet effective WA method to automatically select the robust parameters to conduct WA. 
By assembling the above techniques, we conclude our fast adversarial training method equipped with \textbf{L}ipschitz regularization and
\textbf{A}uto \textbf{W}eight averaging, called FGSM-LAW.
Our main contributions are in three aspects: 
\begin{itemize}
\item We explore existing FAT-related techniques and find efficient and effective fast adversarial training techniques for further research. 

\item We propose a novel regularization approach motivated by the Lipschitz constraint, which can further improve the model robustness for fast adversarial training.

\item We find WA fails to boost the original fast adversarial training and propose a simple yet effective WA method for original fast adversarial training (FGSM-RS) to improve model robustness further. 

\item By assembling these techniques, we conclude our fast adversarial training method, called FGSM-LAW. Compared with state-of-the-art fast adversarial training methods across various network architectures and datasets, FGSM-LAW achieves higher robustness performance with less training costs.
\end{itemize}
    

\section{Related Work}

\subsection{Adversarial Attack Methods}
DNNs are vulnerable to adversarial examples.
Several studies ~\cite{miyato2018virtual,eykholt2018robust,gu2021adversarial,zhang2020attacks,wang2021enhancing,yin2021adv,hu2021naturalistic,duan2021advdrop,hu2022adversarial} focus on generating adversarial examples. In detail,  Goodfellow \emph{et al.}\cite{goodfellow2014explaining} first propose to adopt Fast Gradient Sign Method (FGSM) to attack DNNs, which makes use of the model gradient to generate adversarial examples. Then, a strong adversarial attack method, \emph{i.e.,} Projected Gradient Descent (PGD), is proposed by Madry \emph{et al.}\cite{madry2017towards}. It is an iterative extension of FGSM, which conducts FGSM with the project operation multiple times in a small attack step. To improve the attack performance,  Carlini \emph{et al.}\cite{carlini2017towards} introduce three distance metrics that are used in the previous literature and propose some powerful adversarial attack methods, called C\&W. Andriushchenko \emph{et al.}\cite{andriushchenko2020square} propose a score-based black-box adversarial attack method based on a random search, called Square. Recently, Croce \emph{et al.} \cite{croce2020minimally} propose a fast adaptive boundary attack method that aims to find the minimal adversarial perturbation, called FAB. Moreover, Croce \emph{et al.} \cite{croce2020reliable} propose two powerful adversarial attack methods (APGD-CE and APGD-DLR), which can automatically select the attack step size. They combine the two previous adversarial attack methods (Square and FAB) with their own attack methods, called AutoAttack (AA).

\subsection{Fast Adversarial Training }
\par Even though adversarial training have been proposed to be one of the most effective defense methods, it requires high costs to generate adversarial examples, resulting in its limited practical application. To reduce the training costs, Goodfellow \emph{et al.}\cite{goodfellow2014explaining} propose to adopt FGSM to generate adversarial examples for training, called FGSM-AT. Tram{\`e}r \emph{et al.}\cite{tramer2017ensemble} propose to initialize the clean samples with the Gaussian distribution to conduct FGSM-AT, called R+FGSM. Wong \emph{et al.}\cite{wong2020fast} indicate that both FGSM-AT and R+FGSM easily encounter catastrophic overfitting (CO) and propose to initialize the clean samples with the uniform distribution instead of Gaussian distribution to conduct FGSM-AT, called FGSM-RS. Several fast adversarial training methods are proposed to improve the model robustness from the regularization  perspective. In detail, Anndriushchenko \emph{et al.}\cite{andriushchenko2020understanding} propose a gradient alignment regularization to prevent 
CO and improve the model robustness. Also,  Sriramanan \emph{et al.} \cite{sriramanan2020guided} propose a $\ell_{2}$ distance regularization to improve the quality of adversarial examples. Sriramanan \emph{et al.}  \cite{sriramanan2021towards} propose a Nuclear-Norm regularization to enforce function smoothing. 
Tsiligkaridis  \emph{et al.} \cite{DBLP:conf/cvpr/TsiligkaridisR22} explore the correlation of catastrophic overfitting with low distortion for Frank-Wolfe attacks and propose an adaptive step fast adversarial training, called FW-ADAPT.
Zhang \emph{et al.} \cite{DBLP:conf/icml/ZhangZKHC022} revisit fast adversarial training from bi-level optimization (BLO) and propose a BLO-based fast adversarial training method, called FAST-BAT. Moreover, some research focuses on sample initialization to improve the performance of FGSM-AT. Specifically, Jia \emph{et al.} \cite{jia2022boosting} propose a learnable sample initialization generated by a generative model to boost fast adversarial training. Jia \emph{et al.} \cite{jia2022prior} propose to use the prior adversarial perturbation as the sample initialization to improve robustness.

\section{Techniques to Prevent Overfitting}
\subsection{Rethinking Catastrophic Overfitting}
\label{sec:CO}
Catastrophic Overfitting (CO) indicates the phenomenon that the robust accuracy dramatically and suddenly decreases during the training parse of fast adversarial training, which is first discovered by \cite{wong2020fast}. A series of fast adversarial training variants have been proposed to prevent Catastrophic Overfitting and improve the robustness from two aspects, \emph{i.e.,} sample initialization and regularization. 
Rethink that FGSM attack can be regarded as a closed-form solution to the maximization optimization problem, \emph{i.e.,} $\left.\boldsymbol{\delta}=\arg \max _{\|\boldsymbol{\delta}\|_\infty<\epsilon}\left\langle\nabla_\mathbf{x} \ell(\mathbf{x}, \mathbf{y} ; \boldsymbol{\theta})\right), \boldsymbol{\delta}\right\rangle$. If the loss function is locally linear, the output of $\nabla_\mathbf{x} \ell(\mathbf{x}, \mathbf{y} ; \boldsymbol{\theta})$ is constant within the $\ell_{\infty}$-ball around the input sample $\mathbf{x}$, which can provide the attacker with the optimal adversarial perturbation to attack the model. 
On the contrary, if the loss function is locally nonlinear, as shown in Fig~\ref{fig:CO} (a),  the generated adversarial examples based on one large step FGSM may not be able to reach the local maximum region of the loss function, resulting in low-quality adversarial examples. \cite{jia2022prior, jia2022boosting} have indicated that Catastrophic Overfitting happens when the adversarial example quality becomes worse. 
As shown in Fig~\ref{fig:CO} (b), better initialization can improve the quality of adversarial examples, which can reduce the effect of local nonlinearity to prevent Catastrophic Overfitting. 
 Adding regularization to the loss function is another way of reducing local nonlinearity. In detail, as shown in Fig~\ref{fig:CO} (c), using the regularization can promote the flatness of the loss surface, which mitigates local nonlinearity and meanwhile improves the quality of adversarial examples generated by FGSM. 
Hence, Catastrophic Overfitting is directly related to the quality of the solution to the inner maximization and it is intrinsically caused by the model's local non-linearity. Initialization and regularization improve the quality of the adversarial examples to prevent Catastrophic Overfitting.



\subsection{Sample Initialization}
\label{sec:Initialization}
The sample initialization used in fast adversarial training is quite different. For a given clean sample $\mathbf{x}$ initialized by $\boldsymbol{\eta}$, fast adversarial training methods adopt FGSM to generate the adversarial perturbation $\boldsymbol{\delta}$. It can be defined as:
\begin{equation}
\boldsymbol{\delta}=\Pi_{[-\epsilon, \epsilon]}\left[\boldsymbol{\eta}+\alpha \cdot \operatorname{sign}\left(\nabla_{\mathbf{x}} \mathcal{L}(f(\mathbf{x}+\boldsymbol{\eta} ; \boldsymbol{\theta}), \mathbf{y})\right)\right],
\end{equation}
where $\epsilon$ represents the maximal perturbation, $\alpha$ represents the attack step size, $f(\cdot ;  \boldsymbol{\theta})$ represents the trained model with the weight parameters $ \boldsymbol{\theta}$, $\mathbf{y}$ represents the ground truth label. And the $\mathcal{L}(f(\mathbf{x}+\boldsymbol{\eta} ; \boldsymbol{\theta}), \mathbf{y})$ is the loss function for training. 

\par In detail, the primary FGSM-AT proposed by \cite{goodfellow2014explaining} conduct FGSM with a zero initialization to generate adversarial examples,  \emph{i.e.,} $\boldsymbol{\eta}=0$. Tram{\`e}r \emph{et al.}\cite{tramer2017ensemble} apply 
FGSM with a normal initialization in the half perturbation to generate adversarial examples for training, called \textbf{FGSM-NR}, It can be defined as:
\begin{equation}
\begin{aligned}
\boldsymbol{\delta}= & \Pi_{[-\epsilon, \epsilon]}[\frac{\epsilon}{2} \cdot  \operatorname{Normal}(0,1)\\
& +\alpha \cdot \operatorname{sign}(\nabla_{\mathbf{x}} \mathcal{L}(f(\mathbf{x}+\frac{\epsilon}{2} \cdot  \operatorname{Normal}(0,1) ; \boldsymbol{\theta}), \mathbf{y})) ],
\end{aligned}
\end{equation}
where $\alpha$ is set to $\epsilon /2$ for training. After that, Andriushchenko \emph{et al.} \cite{andriushchenko2020understanding} propose to implement FGSM with a uniform initialization in the whole perturbation to generate adversarial examples for training, called \textbf{FGSM-RS}. It can be defined as:
\begin{equation}
\begin{aligned}
\boldsymbol{\delta}= & \Pi_{[-\epsilon, \epsilon]}[{\epsilon} \cdot  \operatorname{Uniform}(-1,1) \\
& +\alpha \cdot \operatorname{sign}(\nabla_{\mathbf{x}} \mathcal{L}(f(\mathbf{x}+{\epsilon} \cdot \operatorname{Uniform}(-1,1) ; \boldsymbol{\theta}), \mathbf{y})) ],
\end{aligned}
\end{equation}
where $\alpha$ is set to $1.25 \epsilon$ for training. Moreover, in the recent research (\cite{sriramanan2020guided,sriramanan2021towards}), they adopt a Bernoulli initialization in the half perturbation to perform fast adversarial training, called \textbf{FGSM-BR}. It can be defined as:
\begin{equation}
\begin{aligned}
\boldsymbol{\delta}= & \Pi_{[-\epsilon, \epsilon]}[\frac{\epsilon}{2} \cdot  \operatorname{Bernoulli}(-1,1) \\
& +\alpha \cdot \operatorname{sign}(\nabla_{\mathbf{x}} \mathcal{L}(f(\mathbf{x}+\frac{\epsilon}{2} \cdot  \operatorname{Bernoulli}(-1,1) ; \boldsymbol{\theta}), \mathbf{y})) ],
\end{aligned}
\end{equation}
where $\alpha$ is set to $\epsilon$ for training. Also, Zheng \emph{et al.}\cite{zheng2020efficient} propose to adopt the previous adversarial perturbation from the last epoch to initialize the clean sample, called \textbf{ATTA}. It can be defined as:
\begin{equation}
\boldsymbol{\delta}_{E_{t+1}}=\Pi_{[-\epsilon, \epsilon]}\left[\boldsymbol{\delta}_{E_{t},\mathbf{x}}+\alpha \cdot \operatorname{sign}\left(\nabla_{\mathbf{x}} \mathcal{L}(f(\mathbf{x}+\boldsymbol{\delta}_{E_{t},\mathbf{x}} ; \boldsymbol{\theta}), \mathbf{y})\right)\right],
\end{equation}
where $\alpha$ is set to $\epsilon$ for training, $\boldsymbol{\delta}_{E_{t},\mathbf{x}}$ is the generated adversarial perturbation by FGSM attack at the $t$-th epoch on the input sample $\mathbf{x}$. Besides, Jia \emph{et al.} \cite{jia2022prior} adopt series of prior-guided adversarial perturbations as the adversarial initialization which are prior from the previous batch, the previous epoch and the momentum of all previous epochs to perform FGSM, called \textbf{FGSM-PGI}. It can be defined as:
\begin{align}
&\mathbf{g}_{c} =\operatorname{sign}\left(\nabla_{\mathbf{x}} \mathcal{L}(f(\mathbf{x}+\mathbf{\boldsymbol{\eta}}_{E_{t-1}} ; \boldsymbol{\theta}), \mathbf{y})\right), \\
&\mathbf{g}_{E_{t}} =\mu \cdot \mathbf{g}_{E_{t-1}} + \mathbf{g}_{c}, \\
&\mathbf{\boldsymbol{\delta}}_{E_{t}}=\Pi_{[-\epsilon, \epsilon]}\left[\mathbf{\boldsymbol{\eta}}_{E_{t-1}}+\alpha \cdot \mathbf{g}_{c} \right], \label{eq:FSGM_MEP_1}\\
&\mathbf{\boldsymbol{\eta}}_{E_{t}} =\Pi_{[-\epsilon, \epsilon]}\left[\mathbf{\boldsymbol{\eta}}_{E_{t-1}}+\alpha \cdot \operatorname{sign}(\mathbf{g}_{E_{t}}) \right]. \label{eq:FSGM_MEP_2}
\end{align}
where $\alpha$ is set to $\epsilon$ for training, $\mathbf{g}_c$ is regarded as the signed gradient and $\mathbf{g}_{E_{t}}$ is the signed gradient momentum in the $t$-th epoch. Moreover, Jia \emph{et al.} \cite{jia2022boosting} propose to adopt a generative model to generate the sample initialization for training, called \textbf{FGSM-SDI}. It can be defined as:
\begin{equation}
\begin{aligned}
\boldsymbol{\delta}= & \Pi_{[-\epsilon, \epsilon]}[\epsilon \cdot g\left(\mathbf{x}, \mathbf{x}_{g} ; \mathbf{w}\right) \\
& +\alpha \cdot \operatorname{sign}(\nabla_{\mathbf{x}} \mathcal{L}(f(\mathbf{x}+\epsilon \cdot g\left(\mathbf{x}, \mathbf{x}_{g} ; \mathbf{w}\right) ; \boldsymbol{\theta}), \mathbf{y}))],
\end{aligned}
\end{equation}
where $\alpha$ is set to $\epsilon$ for training, $\mathbf{x}_{g}$ represents gradient information of $\mathbf{x}$ on the trained model and $g\left(\cdot , \cdot ; \mathbf{w}\right)$ represents the generative model with the weight parameters $\mathbf{w}$.

\par We summarize existing initialization methods used in fast adversarial training and divide them into three categories: \textbf{a) random-based initialization, b) prior-based initialization and c) learning-based initialization}. 

\par \textbf{Random-based Initialization} 
Most fast adversarial training methods \cite{goodfellow2014explaining,tramer2017ensemble,sriramanan2021towards} adopt random-based initialization methods to conduct AT. They always use a random sample-independent distribution to initialize the sample for the adversarial example generation, \emph{i.e.,}
$\boldsymbol{\eta} \sim \operatorname{Random}(\epsilon)$, where $\operatorname{Random}(\epsilon)$
represents a random distribution related to the maximum perturbation strength $\epsilon$. 
In detail, Goodfellow \emph{et al.}\cite{goodfellow2014explaining} conduct FGSM with a zero initialization to generate adversarial examples,  \emph{i.e.,} $\boldsymbol{\eta}=0$. 
Tram{\`e}r \emph{et al.}\cite{tramer2017ensemble} apply FGSM with a normal initialization in the half perturbation to generate adversarial examples for training, \emph{i.e.,} $\boldsymbol{\eta}=\frac{\epsilon}{2} \cdot \operatorname{Normal}(0,1)$, called FGSM-NR. 
Wong \emph{et al.}\cite{wong2020fast} propose to implement FGSM with a uniform initialization in the whole perturbation to generate adversarial examples for training, \emph{i.e.,} $\boldsymbol{\eta}={\epsilon} \cdot  \operatorname{Uniform}(-1,1)$, called FGSM-RS. Moreover, in the recent researchs \cite{sriramanan2020guided,sriramanan2021towards}, they adopt a Bernoulli initialization in the half perturbation to perform fast adversarial training, \emph{i.e.,} $\boldsymbol{\eta}=\frac{\epsilon}{2} \cdot  \operatorname{Bernoulli}(-1,1)$, called FGSM-BR.

\par \textbf{Prior-based Initialization} Several works~\cite{zheng2020efficient,jia2022prior} also propose to use the previous adversarial perturbations as the sample initialization, which can be called  prior-based initialization, \emph{i.e.,}
$\boldsymbol{\eta}\sim\operatorname{Prior}(\epsilon,\mathbf{x}),$
where $ \operatorname{Prior}(\epsilon, \cdot)$
represents the function related to the maximum perturbation strength $\epsilon$ and the sample $\mathbf{x}$, \emph{i.e.,} sample-dependent initialization. Zheng \emph{et al.} \cite{zheng2020efficient} propose to adopt the previous adversarial perturbation from the last epoch to initialize the clean sample, \emph{i.e.,} $\boldsymbol{\eta}=\boldsymbol{\delta}_{E}$, called ATTA. And Jia \emph{et al.} \cite{jia2022prior} propose to use some prior-guided adversarial perturbations as the adversarial initialization which is prior from the previous epoch and the momentum of all previous epochs to perform FGSM , \emph{i.e.,} $\boldsymbol{\eta}=\boldsymbol{\delta}_{P}$, called FGSM-PGI. Note that in this section, for a fair comparison,
we adopt the FGSM-PGI without the proposed regularization in their original paper to conduct experiments.

\par \textbf{Learning-based Initialization}
Jia \emph{et al.} \cite{jia2022boosting} propose to adopt a sample-dependent learnable initialization to conduct AT. In detail, they adopt a generative model to generate the sample initialization for training, called FGSM-SDI, \emph{i.e.,}
$\boldsymbol{\eta}=\alpha \cdot g\left(\mathbf{x}, \nabla_{x} {\mathcal{L}(\mathbf{x}, \mathbf{y})} ; \mathbf{w}\right),$
where $\alpha$ is set to $\epsilon$ for training,
and $g\left(\cdot , \cdot ; \mathbf{w}\right)$ represents the generative model with the weight parameters $\mathbf{w}$.

\begin{figure*}[t]
\begin{center}
   \includegraphics[width=1\linewidth]{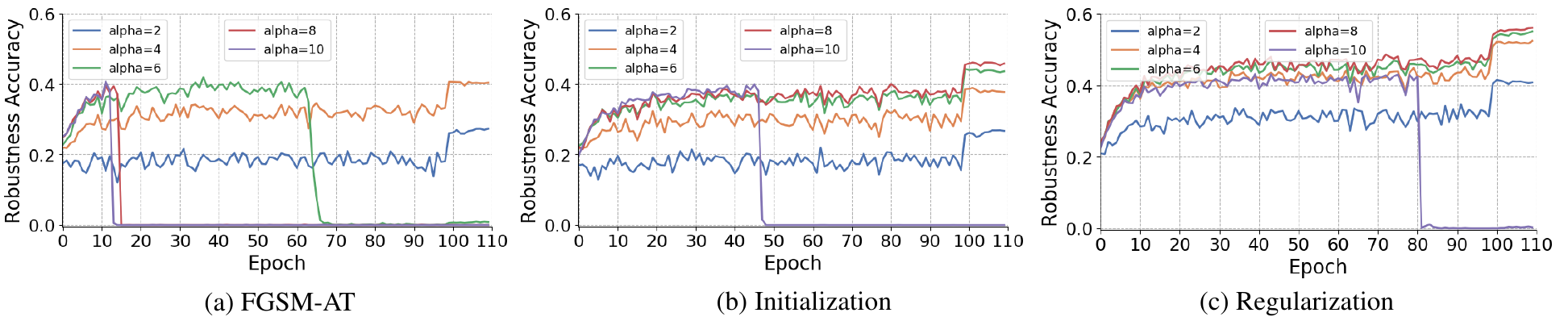}
\end{center}
\vspace{-4mm}
\caption{The robustness accuracy under PGD-10 of FGSM-AT, FGSM with Bernoulli random initialization and FGSM with the regularization of different step sizes on the CIFAR-10 during the training phase. 
}
\label{fig:Fig_Initialization}
\vspace{-4mm}
\end{figure*}
\begin{table}[t]
\centering
\caption{ Comparisons of clean and robust accuracy (\%) and training time on the CIFAR-10 database using ResNet18 with different initialization techniques. Number in bold indicates the best. 
}
\label{table:cifar10_init}
\scalebox{0.75}{
\begin{tabular}{ccccccc}
\toprule
Method   & Type           & \begin{tabular}[c]{@{}c@{}}Clean\end{tabular} & \begin{tabular}[c]{@{}c@{}}PGD-50\end{tabular} & \begin{tabular}[c]{@{}c@{}}C\&W\end{tabular} & \begin{tabular}[c]{@{}c@{}}AA\end{tabular} & \begin{tabular}[c]{@{}c@{}}Training\\ Time (min)\end{tabular} \\ \midrule
FGSM-NR~\cite{tramer2017ensemble}  & Random-based   & 86.41                                               & 43.99                                                & 45.4                                               &41.29                                             & 51  \\ \midrule
FGSM-RS~\cite{wong2020fast}  & Random-based   & 86.65                                               & 44.12                                               & 45.65                                              & 42.2                                             & 51   \\ \midrule
FGSM-BR~\cite{sriramanan2020guided,sriramanan2021towards}  & Random-based   & \textbf{86.82}                                               & 44.23                                                & 45.89                                              & 42.37                                            & 51   \\ \midrule
ATTA~\cite{zheng2020efficient}     & Prior-based    & 86.30                                               & 44.97                                                & 43.8                                               & 42.84                                            & 51   \\ \midrule
FGSM-PGI~\cite{jia2022prior} & Prior-based    & 86.61                                               & 45.69                                                & 44.8                                                & 43.26                                            & 51   \\ \midrule
FGSM-SDI~\cite{jia2022boosting} & Learning-based & 85.25                                               & \textbf{51.79}                                               & \textbf{50.29}                                             & \textbf{47.91}                                           & 83  \\ \bottomrule
\end{tabular}
}
\end{table}
\par \textbf{Discussion} Previous works adopt different sample initialization for AT to prevent Catastrophic Overfitting and improve the model robustness. As shown in Fig.~\ref{fig:CO}, using a zero initialization, FGSM with a large step size obtains a worse solution to the inner maximization problem. But, using sample initialization, FGSM can achieve a better solution to the inner maximization problem. The quality of the solutions to the inner maximization problem is related to the Catastrophic Overfitting. As shown in Fig.~\ref{fig:Fig_Initialization}, FGSM-AT with a small step size (e.g, $\alpha=4/255$) does not meet Catastrophic Overfitting and the attack success rate and model robustness are directly related in fast adversarial training. It indicates that when the quality of the inner maximization problem becomes worse, Catastrophic Overfitting could happen. 
Using a zero initialization, FGSM with a larger step size more easily meets Catastrophic Overfitting. Using sample initialization can prevent Catastrophic Overfitting when FGSM with a large step size.
We summarize the existing initialization methods and make some meaningful findings based on the quality of the solutions to the inner maximization problem. 
First, sample-dependent initialization methods (prior-based and learning-based initialization)  can achieve better robustness performance than sample-independent initialization methods (random-based initialization). The learning-based initialization achieves the best robustness performance. Second, surprisingly, random-based initialization methods with the appropriate step size  prevent Catastrophic Overfitting and achieve approximately the same robustness improvement performance, which has not been explored in the previous works. 
We adopt ResNet18 for AT with different initialization to conduct experiments on CIFAR10.
The result is shown in Table~\ref{table:cifar10_init}, which demonstrates the correctness of our findings. Compared with the previous random initialization, 
the advanced ATTA and FGSM-PGI need more memory to store the historical adversarial perturbations, which could require more memory costs. And FGSM-SDI needs more time consumption to train an extra generator. We use random initialization based on the Bernoulli initialization and  the appropriate step size to conduct fast adversarial training since it can achieve comparable robustness with less training costs.

\subsection{Regularization}
Regularization used in AT can be divided into two categories: a) regularization worked in the inner maximization and b) regularization worked in the outer minimization. The former is related to the inner maximization to improve the quality of the adversarial examples. The latter is related to the outer minimization to improve the model robustness. 
Existing fast adversarial training variants adopt the regularization in min-max optimization or only in minimization optimization. It can be defined as:
\begin{equation}
\begin{aligned}
&\boldsymbol{\delta}^{*} = \text{argmax}_{\boldsymbol{\delta} \in \Omega} [\mathcal{L}(f(\mathbf{x}+\boldsymbol{\delta} ; \boldsymbol{\theta}), \mathbf{y}) + \gamma \cdot R(\mathbf{x}, \mathbf{x}+\boldsymbol{\delta};\boldsymbol{\theta}) ] \\
& \min _{\boldsymbol{\theta}} \mathbb{E}_{(\mathbf{x}, \mathbf{y}) \sim \mathcal{D}}\left[  \mathcal{L}(f(\mathbf{x}+\boldsymbol{\delta}^{*} ; \boldsymbol{\theta}), \mathbf{y}) + \mu \cdot R(\mathbf{x}, \mathbf{x}+\boldsymbol{\delta}^{*};\boldsymbol{\theta})  \right],
\end{aligned}
\end{equation}
where $\gamma, \mu \in [0,1]$ control the use of regularization and $ R(\mathbf{x}, \mathbf{x}+\boldsymbol{\delta} ;\boldsymbol{\theta})$ represents the regularization.
\par \textbf{Regularization only in minimization optimization}
 Andriushchenko \emph{et al.} \cite{andriushchenko2020understanding} propose a gradient alignment regularization to measure the local linearity.  
They adopt the regularization only in the minimization optimization, \emph{i.e.,} $\gamma=0$ and $\mu=1$.
It can be defined as: 
\begin{equation}
\begin{aligned}
& R\left(\mathbf{x}, \mathbf{x}+\boldsymbol{\delta} ; \boldsymbol{\theta}\right) \\
& =1-\cos \left(\nabla_{\mathbf{x}} \mathcal{L}(f(\mathbf{x} ; \boldsymbol{\theta}), \mathbf{y}), \nabla_{\mathbf{x}} \mathcal{L}(f(\mathbf{x}+\boldsymbol{\eta} ; \boldsymbol{\theta}), \mathbf{y})\right),
\end{aligned}
\end{equation}
where $\boldsymbol{\eta}$ represents the sample initialization. Although the FGSM-GA achieves model robustness improvement, 
it requires extra training time to compute the alignment regularization on the gradient.

\par \textbf{Regularization in min-max optimization}
Sriramanan \emph{et al.} \cite{sriramanan2020guided} propose a more effective guided regularization for improved optimization and function smoothing. They adopt regularization in min-max optimization. 
It works in the inner maximization to generate adversarial examples and the outer minimization to improve the model robustness,\emph{i.e.,} $\gamma=1$ and $\mu=1$.
It can be defined as:
\begin{equation}
 R(\mathbf{x}, \mathbf{x}+\boldsymbol{\delta}; \boldsymbol{\theta})=\lambda \cdot \left\|f(\mathbf{x}+\boldsymbol{\delta} ; \boldsymbol{\theta}) -f(\mathbf{x} ; \boldsymbol{\theta})\right\|_2^2,
\end{equation}
where 
$\lambda$ represents the a hyper-parameter that determines the smoothness of the loss surface. Sriramanan \emph{et al.} \cite{sriramanan2021towards} propose a Nuclear-Norm regularization for enforcing function smoothing in min-max optimization, \emph{i.e.,} $\gamma=1$ and $\mu=1$. 
It can be defined as:
\begin{equation}
 R(\mathbf{x}, \mathbf{x}+\boldsymbol{\delta}; \boldsymbol{\theta}) = \lambda \cdot \left\|f(\mathbf{x}+\boldsymbol{\delta} ; \boldsymbol{\theta}) -f(\mathbf{x} ; \boldsymbol{\theta})\right\|_*,
\end{equation}
where $\|\cdot\|_*$ represents the Nuclear Norm which is the sum of the singular values. 

\par \textbf{Discussion} 
We conclude that the regularization in min-max optimization achieves better robustness. 
It greatly improves model robustness. 
In inner maximization, the regularization loss is used to find the least smooth domain of the loss surface. In outer minimization, the regularization is used to improve the local smoothness of the loss surface. 
Although the regularization is effective to improve the model robustness, it also requires a computation overhead. On the other hand, adversarial examples almost exist in the least smooth domain of the loss surface \cite{rebuffi2021fixing,wu2020adversarial} which has the same effect as inner maximization in regularization. Based on this, we argue that it is not necessary to use regularization in the inner maximization. 
To verify this opinion, we conduct experiments on CIFAR-10 with ResNet18 comparing regularization in the min-max optimization and regularization solely in the minimization. We adopt FGSM with BR initialization as the baseline. 
The results are shown in Table~\ref{table:cifar10_regularization}.  Compared with fast adversarial training with the regularization in min-max optimization, 
Fast adversarial training with the regularization only in the minimization achieves better performance under PGD attack scenarios and comparable robustness performance under C\&W and AA attack scenarios. The regularization only in the minimization requires less training time. 

\begin{table}[t]
\centering
\caption{ Comparisons of clean and robust accuracy (\%) and training time on the CIFAR-10 database using ResNet18 with different regularization methods. Number in bold indicates the best. 
}
\label{table:cifar10_regularization}
\scalebox{0.75}{
\begin{tabular}{cccccc}
\toprule
Method        & Clean  & PGD-50 & C\&W  & AA    & \begin{tabular}[c]{@{}c@{}}Training\\ Time (min)\end{tabular} \\ \midrule \midrule
Baseline    & 86.82   & 44.23  & 45.89 & 42.37  & 51   \\ \midrule \midrule
Guided Regularization(min-max)  & 83.33  & 50.96  & 48.64 & 46.56 &  102    \\ \midrule
Guided Regularization(min)      & 84.20   & 51.26  & 47.87 & 45.74 &  73  \\ \midrule
Nuclear-Norm Regularization(min-max) & 84.9    & 49.79  & 49.14 & 47.36 & 102   \\ \midrule
Nuclear-Norm Regularization(min)     & \textbf{84.94}    & 50.78  & 48.83 & 47.06 & 73 \\ \midrule 
Ours(min)     & 82.97  & \textbf{54.29}  & \textbf{50.41} & \textbf{48.05} & 73 \\ \bottomrule
\end{tabular}}
\vspace{-4mm} 
\end{table}

\subsection{The Proposed Regularization } 
\label{sec:regularization}
\begin{figure*}[t]
\begin{center}
   \includegraphics[width=1.0\linewidth]{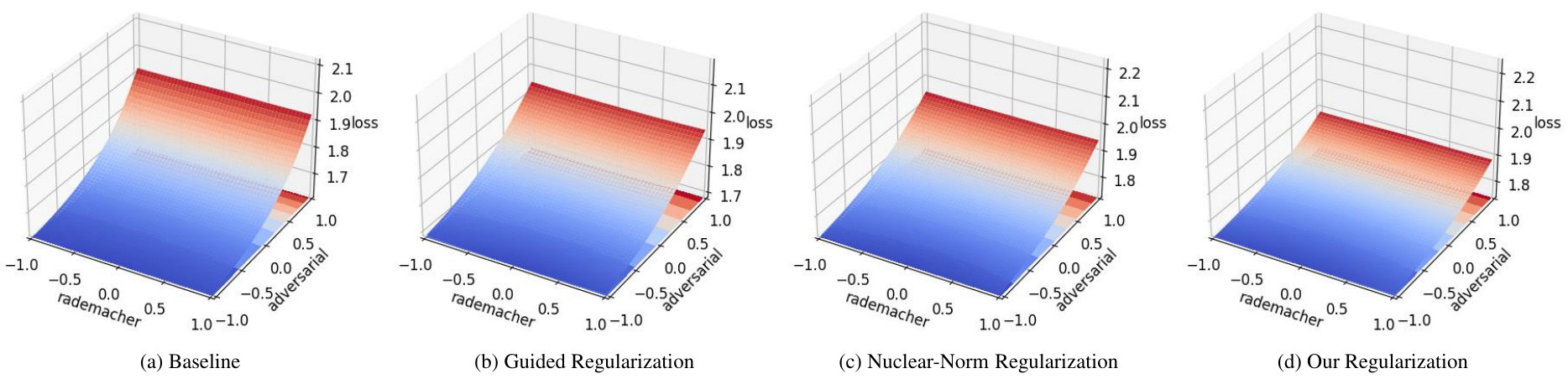}
\end{center}

\caption{Visualization of the loss landscape on CIFAR10 of different regularization methods, \emph{i.e.,} the baseline FGSM-BR, FGSM-BR with the guided regularization(min-max),  FGSM-BR with the nuclear-norm regularization(min-max) and FGSM-BR with our proposed regularization. We randomly select 1000 test images for CIFAR-10 testing dataset and plot the cross-entropy loss along a spatial variation composed of two directions, \emph{i.e.,} a Rademacher (random) direction $r_{1}$ and an adversarial direction $r_{2}$. The Rademacher (random) direction is calculated as:  $r_{1} \sim \operatorname{Rademacher}(\eta)$, where $\eta$ is $8/255$. The adversarial direction is calculated as: $r_{2}=\eta \operatorname{sign}(\nabla_{x} f(\mathbf{x}; \boldsymbol{\theta}))$. Note the same adversarial attack  (PGD-50) are conducted on the same images for the visualization.}

\label{fig:Fig_loss_landscape}

\end{figure*}

Previous works \cite{weng2018evaluating,wu2021wider} have demonstrated that the robustness is related to the local Lipschitzness, \emph{i.e.,} smaller local Lipschitzness leads to stronger robustness. We propose a novel regularization approach motivated by the Lipschitz constraint. It can be defined as:
\begin{equation}
\label{eq:prop_regular}
\begin{aligned}
\frac{\lambda}{{\left\|\boldsymbol{\delta} -\boldsymbol{\eta} \right\|_2^2}} \cdot & (\left\|f(\mathbf{x}+\boldsymbol{\delta} ; \boldsymbol{\theta}) -f(\mathbf{x}+\boldsymbol{\eta} ; \boldsymbol{\theta})\right\|_2^2 \\
& +\left\|f_{F}(\mathbf{x}+\boldsymbol{\delta} ; \boldsymbol{\theta}) -f_{F}(\mathbf{x}+\boldsymbol{\eta} ; \boldsymbol{\theta})\right\|_2^2),
\end{aligned}
\end{equation} 
where $f_{F}(\cdot ; \boldsymbol{\theta})$ represents the feature output of $f(\cdot ; \boldsymbol{\theta})$ and $\lambda$ is the hyper-parameter. We also compared the proposed regularization with the previous regularization methods, the results are shown in Table~\ref{table:cifar10_regularization}.
The proposed regularization achieves the best performance under all attack scenarios. We also visualize the loss landscape of each regularization methods. As shown in Fig.~\ref{fig:Fig_loss_landscape}, it can be observed that 
using the regularization methods can make the cross-entropy loss more linear in the adversarial direction. Our proposed regularization method achieves the most linear cross-entropy loss in the adversarial direction. It indicates 
that our method can improve the local linearity of the trained model better. 
Moreover, We provide proof to prove the effectiveness of \textbf{the proposed regularization} in Supplementary Material.

\section{Techniques to Improve Model Robustness}

\begin{table}[t]
\centering
\caption{ Test robustness (\%) on the CIFAR-10 database using ResNet18 with different data augmentation techniques. Number in bold indicates the best. 
}
\label{table:cifar10_augmentation}
\scalebox{0.8}{
\begin{tabular}{ccccccc}
\toprule
Method      & Clean & PGD-10 & PGD-20 & PGD-50 & C\&W  & AA    \\ \midrule
Baseline    & 86.82 & 46.62  & 44.76  & 44.23  & 45.89 & 42.37     \\ \midrule \midrule
FGSM-BR+Mixup       & \textbf{86.94} & 46.78  & 44.93  & 44.41  & 45.52 & 42.34   \\ \midrule
FGSM-BR+Cutout      & 84.93 & \textbf{48.31}  & \textbf{46.9}   & \textbf{46.43}  & \textbf{46.91} & \textbf{43.79}     \\ \midrule
FGSM-BR+AutoAugment & 85.79 & 48.97  & 47.24  & 46.57  & 46.06 & 42.25  \\ \midrule
FGSM-BR+CutMix      & 81.12 & 50.45  & 49.28  & 48.95  & 45.36 & 42.87     \\ \bottomrule
\end{tabular}}
\end{table}

\subsection{Data Augmentation}
\label{sec:Augmentation}
\begin{figure}[t]
\begin{center}
   \includegraphics[width=0.85\linewidth]{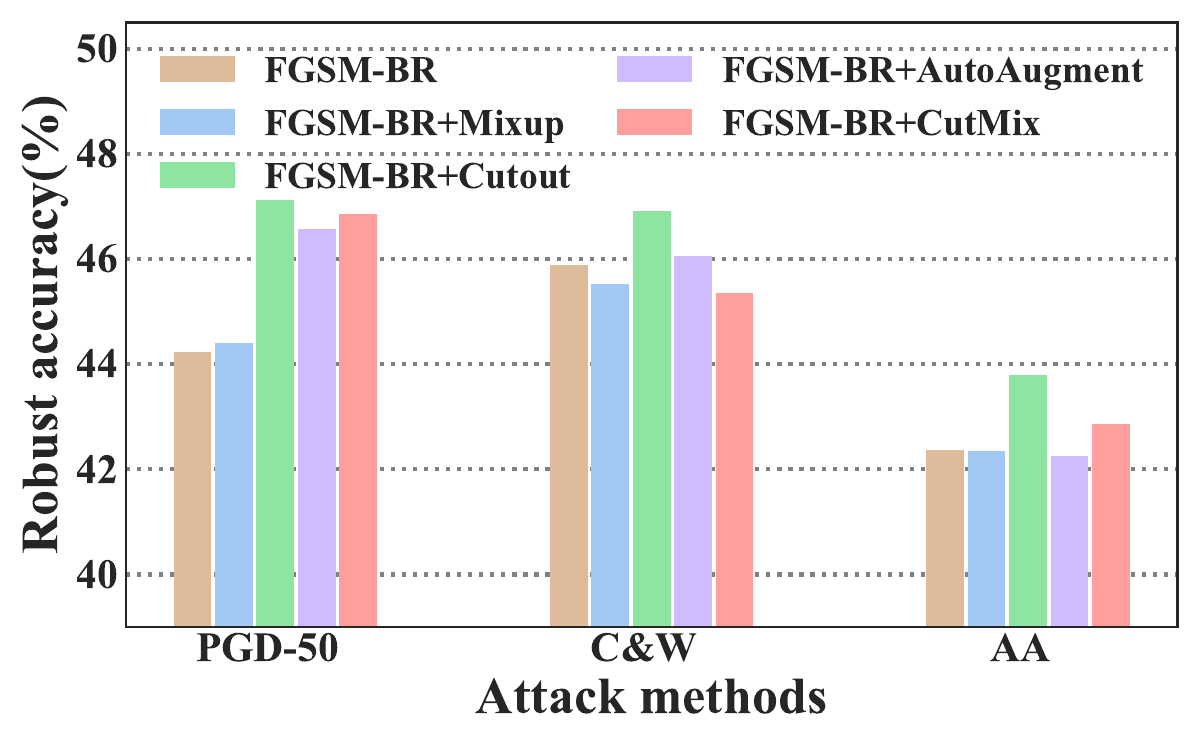}
\end{center}

\caption{The adversarial robustness performance  of FAT methods with different data augmentation methods on CIFAR-10 under several attack scenarios. }

\label{fig:aug}
\end{figure}

Several works ~\cite{gowal2020uncovering,rebuffi2021fixing} have explored the impact of more sophisticated data augmentation techniques on the robustness of PGD-based AT. However, the impact of the data augmentation techniques on model robustness of FGSM-based AT has not been studied before. In this paper, we introduce the sophisticated  data augmentation techniques (Cutout~\cite{devries2017improved}, Mixup~\cite{zhang2017mixup}, CutMix~\cite{yun2019cutmix}, and AutoAugment~\cite{cubuk2018autoaugment}) into existing fast adversarial training techniques and explore the impact of them on model robustness of fast adversarial training methods. 
In detail, we use FGSM-BR as the baseline and compare it with FGSM-BR with data augmentation techniques. 
The results are shown in Fig.~\ref{fig:aug}. 
In detail, we adopt the PGD-10, PGD-20, PGD-50, C\&W and AA to evaluate the FGSM-BR with different data augmentation methods. The result is shown in Table~\ref{table:cifar10_augmentation}. It is clear that using Cutout, Mixup, CutMix and AutoAugment can improve model robustness. FGSM-BR combined with Cutout achieves the best robustness performance in all attack scenarios. Moreover, Cutout also achieves comparable clean accuracy on the clean images to the vanilla fast adversarial training (FGSM-BR).
Hence, it can be observed that combined with data augmentation, fast adversarial training can achieve better robustness performance in all the attack scenarios. Particularly, under the PGD-10 attack, Cutout improves the performance by about 1.5\%. Compared with the vanilla FGSM-BR, using data augmentation methods can improve the quality of the solution to the inner maximization problem, \emph{i.e.,} the higher attack success rate. Unlike PGD-based AT~\cite{rebuffi2021fixing}, using data augmentation techniques can improve the robustness of FGSM-based adversarial training. 

\subsection{Model Weight Averaging}

As a technique to improve the model generalization, model weight averaging (WA) ~\cite{izmailov2018averaging} obtains a WA model by computing exponentially weighted moving averages of model parameters of each training iteration. Formally, given a decay rate $\tau$ and the model parameters $\boldsymbol{\theta}$ of an iteration, the WA model is updated with 
$\tilde{\boldsymbol{\theta}} \leftarrow \tau \cdot \tilde{\boldsymbol{\theta}}+(1-\tau) \cdot \boldsymbol{\theta}$. The obtained WA model $\tilde{\boldsymbol{\theta}}$ at the end of training often generalizes better to the unseen test data.
Recent work \cite{chen2020robust, wu2020adversarial} has indicated that the WA model shows higher robustness by achieving a flatter adversarial loss landscape. In particular, Wong \emph{et al.}\cite{wang2022self} show that the WA model obtained in SAT also shows higher adversarial robustness. However, the impact of WA on the fast adversarial training has not been explored in previous works. We make the first exploration by combining the FGSM-RS with the vanilla EM~\cite{rebuffi2021fixing} and SE-EMA~\cite{wang2022self}. 
As shown in the Table~\ref{table:wa}, simply applying the existing WA to original fast adversarial training (FGSM-RS) does not improve the robustness.

\subsection{The Proposed Model Weight Averaging}
In this work, we attribute the failure of WA on original fast adversarial training (FGSM-RS) to the weak quality of adversarial examples generated by FGSM. The existing model weight averaging (WA) will accumulate model parameters of all iterations regardless of the quality of adversarial examples. However, as discussed in Sec.~\ref{sec:CO}, the gradient directions can be undesired when the quality of adversarial examples in fast adversarial training is low. The model parameter updates with the inaccurate gradients can be accumulated in WA model.

To overcome the shortcoming of the existing WA methods, we propose to exclude some training iterations from the WA model update trajectory. Namely, the WA model will not update in an iteration as long as the quality of adversarial examples in the iteration is low. Specifically, we propose a simple yet effective metric to automatically evaluate the quality of adversarial examples. The metric will be used to determine whether the WA model will be updated in the current training iteration. Concretely, the metric is defined as the ratio of the robust accuracy on adversarial examples to the corresponding clean accuracy to evaluate the adversarial example quality.
Formally, it can be formulated as:
\begin{equation}
\Delta=\frac{Acc(f(\mathbf{x}_{b}+\boldsymbol{\delta}_{b} ; \boldsymbol{\theta}),\mathbf{y}_{b})}{Acc(f(\mathbf{x}_{b} ; \boldsymbol{\theta}),\mathbf{y}_{b})},
\end{equation}
where $\mathbf{x}_{b}$ is the input samples within a batch, $\boldsymbol{\delta}_{b}$ is the generated perturbations within a batch, $\mathbf{y}_{b}$ is the corresponding labels within a batch and $Acc(\cdot)$ is the accuracy of trained model on training samples. When the proposed metric $\Delta$ is above a certain threshold $T$, the trained parameters $\boldsymbol{\theta}$ is added to the average model parameters $\tilde{\boldsymbol{\theta}}$, \emph{i.e,}
$\tilde{\boldsymbol{\theta}} \leftarrow \tau \cdot \tilde{\boldsymbol{\theta}}+(1-\tau) \cdot \kappa \cdot \boldsymbol{\theta},$
where $\kappa= \mathbf{1} (\Delta <= T)$.
In summary, our proposed method improves EMA by automatically selecting iterations to update WA model, called Auto-EMA. We conduct original FGSM-RS with our Auto-EMA and compare it with Vanilla EMA and SE-EMA. The result is shown in Table~\ref{table:wa}. It can be observed that EMA and SE-EMA also meets CO but the proposed Auto-EMA can prevent CO and achieves the best robustness. 

\begin{table}[t]
\centering
\caption{ Comparisons of clean and robust accuracy (\%) and training time on the CIFAR-10 database using ResNet18 with different model weight averaging techniques under $\ell_{\infty}= 8/225$. Number in bold indicates the best. 
}
\label{table:wa}
\scalebox{0.75}{
\begin{tabular}{cccccccc}
\toprule
Method      & Clean          & PGD-10         & PGD-20         & PGD-50        & C\&W           & AA    & \begin{tabular}[c]{@{}c@{}}Training\\ Time (min)\end{tabular}      \\ \midrule
FGSM-RS~\cite{wong2020fast}   & 83.82 & 0.0 & 0.0  & 0.0  & 0.0& 0.0  & 51 \\ \midrule \midrule
Vanilla EMA~\cite{rebuffi2021fixing}  & \textbf{84.9} & 0.0           & 0.0          & 0.0          & 0.0          & 0.0  & 52 \\ \midrule
SE-EMA~\cite{wang2022self}      & 84.8          & 0.0          & 0.0          & 0.0        & 0.0          & 0.0 & 52 \\ \midrule
Auto-EMA(Ours)        & 83.62          & \textbf{51.26} & \textbf{50.21} & \textbf{49.9} & \textbf{49.42} &  \textbf{46.53}     & 52 \\ \bottomrule
\end{tabular}}
\vspace{-2mm}
\end{table}
\begin{table}[h]
\centering
\caption{ Test robustness (\%) on the CIFAR-10 database using ResNet18 with equipped with Lipschitz regularization with different hyper-parameter $\lambda$ . Number in bold indicates the best. 
 }
\label{table:HP_regular}
\scalebox{0.75}{
\begin{tabular}{cccccccc}
\toprule
 Hyper-Parameter & Clean          & PGD-10         & PGD-20         & PGD-50         & C\&W           & AA             & \begin{tabular}[c]{@{}c@{}}Training\\ Time (min)\end{tabular}  \\ \midrule
Baseline    & \textbf{86.82} & 46.62  & 44.76  & 44.23  & 45.89 & 42.37  & 51    \\ \midrule \midrule
$\lambda=8$      & 81.61   & 56.31         & 55.56          & 55.48         & 50.83         & 48.77           & 81 \\ \midrule
$\lambda=10$     & 81.91          & 56.73          & 56.03          & 55.9          & 51.32          & 49.12         & 81 \\ \midrule
$\lambda=12$     & 81.34          & \textbf{57.07}          & \textbf{56.57}         & \textbf{56.36}         & \textbf{51.4}          & \textbf{49.22}          & 81 \\ \midrule
$\lambda=14$    & 81.40         & 57.05 & 56.31 & 56.11 & 51.08 & 48.71 & 81 \\ \midrule
$\lambda=16$     & 81.25          & 56.98         & 56.25          & 56.15          & 51.03          & 48.96          & 81  \\ \midrule
$\lambda=18$     & 80.93          & 57.06           & 56.45          & 56.31          & 51.25          & 49.10         & 81  \\ \bottomrule
\end{tabular}
}
\end{table}

\begin{table}[h]
\centering
\caption{ Test robustness (\%) on the CIFAR-10 database using ResNet18 equipped with auto weight averaging with diffrerent hyper-parameters $\Delta$. Number in bold indicates the best. 
 }
\label{table:HP_WA}
\scalebox{0.75}{
\begin{tabular}{cccccccc}
\toprule
Hyper-Parameter & Clean & PGD-10         & PGD-20        & PGD-50         & C\&W           & AA             & \begin{tabular}[c]{@{}c@{}}Training\\ Time (min)\end{tabular} \\ \midrule
Baseline    & \textbf{86.82} & 46.62  & 44.76  & 44.23  & 45.89 & 42.37  & 51    \\ \midrule \midrule
$\Delta=0.8$    & 80.40 & 56.97 & 56.4 & 56.26 & 51.35         & 48.98          & 81    \\ \midrule
$\Delta=0.82$   & 81.34 & \textbf{57.07}          & \textbf{56.57}          & \textbf{56.36}          & \textbf{51.4}          & \textbf{49.22}          & 81   \\ \midrule
$\Delta=0.84$    & 82.91 & 56.12          & 55.27         & 55.12          & 51.17          & 48.63         & 81    \\ \midrule
$\Delta=0.86$    & 82.99 & 56.12           & 55.20         & 55.0         & 51.0          & 48.67          & 81    \\ \midrule
$\Delta=0.88$   & 83.08 & 56.03          & 54.97         & 54.95           & 50.75 & 48.23 & 81    \\ \midrule
$\Delta=0.9$   & 83.44 & 55.52          & 54.64         & 54.46         & 50.8          & 48.32         & 81    \\ \bottomrule
\end{tabular}
}
\end{table}

\section{Experiments}
\label{sec:ImageNet}
To evaluate the proposed FGSM-LAW, We adopt several datasets, \emph{i.e,} CIFAR-10~\cite{krizhevsky2009learning}, CIFAR-100~\cite{krizhevsky2009learning}, Tiny ImageNet~\cite{deng2009imagenet} and ImageNet~\cite{deng2009imagenet}, which are widely used to evaluate the model robustness. 

\subsection{Default Training Setting}
\label{sec:cyclic}
Following the default settings of AT \cite{pang2020bag,jia2022boosting}, we apply the same training hyper-parameters to conduct experiments. All models with ReLU activation function are trained for 110 epochs with the batch size 128. We use SGD momentum optimizer with the weight
decay of $5 \times 10^{-4}$.  The learning rate initialized to 0.1 decays with a factor of 0.1 at  100 and 105 epochs. All experiments are conducted under the $\ell_{\infty}$ distance metric with the maximal perturbation of $8 / 255$. We adopt PGD attack of 50-steps (\textbf{PGD-50}), C\&W attack of 20 steps (\textbf{C\&W}) and AutoAttack (\textbf{AA}) to evaluate the trained models. Note that we report the results of the last checkpoint, and we also report the the results of the best checkpoint with the best accuracy under PGD-10 attack in the Supplementary Material.
All experiments are conducted 
on a single NVIDIA Tesla V100 for a fair comparison. 

\begin{table*}[t]
\centering
\centering
\caption{ Comparisons of clean and robust accuracy (\%) and training time on the CIFAR-10, CIFAR-100 and Tiny ImageNet databases using different adversarial training methods under $\ell_{\infty}= 8/225$. Number in bold indicates the best. 
}
\label{table:experiments}
\scalebox{0.85}{
\begin{tabular}{@{}c|cccc|c|cccc|c|cccc|c@{}}
\toprule
Dataset         & \multicolumn{4}{c|}{CIFAR-10}                                                                                                    & \multirow{2}{*}{\begin{tabular}[c]{@{}c@{}}Training \\ Time(min)\end{tabular}} & \multicolumn{4}{c|}{CIFAR-100}                                                                                                   & \multirow{2}{*}{\begin{tabular}[c]{@{}c@{}}Training \\ Time(min)\end{tabular}} & \multicolumn{4}{c|}{Tiny ImageNet}                                                                                               & \multirow{2}{*}{\begin{tabular}[c]{@{}c@{}}Training \\ Time(min)\end{tabular}} \\ \cmidrule(r){1-5} \cmidrule(lr){7-10} \cmidrule(lr){12-15}
Method          & \multicolumn{1}{c|}{Clean}          & \multicolumn{1}{c|}{PGD-50}         & \multicolumn{1}{c|}{C\&W}           & AA             &                                                                                & \multicolumn{1}{c|}{Clean}          & \multicolumn{1}{c|}{PGD-50}         & \multicolumn{1}{c|}{C\&W}           & AA             &                                                                                & \multicolumn{1}{c|}{Clean}          & \multicolumn{1}{c|}{PGD-50}         & \multicolumn{1}{c|}{C\&W}           & AA             &                                                                                \\ \midrule
PGD-AT~\cite{rice2020overfitting}          & \multicolumn{1}{c|}{82.65}          & \multicolumn{1}{c|}{52.27}          & \multicolumn{1}{c|}{51.28}          & 48.93          & 265                                                                         & \multicolumn{1}{c|}{57.5}           & \multicolumn{1}{c|}{28.9}           & \multicolumn{1}{c|}{27.6}           & 25.58          & 284                                                                          & \multicolumn{1}{c|}{45.28}          & \multicolumn{1}{c|}{15.4}           & \multicolumn{1}{c|}{14.28}          & 12.84          & 1833                                                                          \\ \midrule
FGSM-RS~\cite{wong2020fast}        & \multicolumn{1}{c|}{83.82}          & \multicolumn{1}{c|}{0.02}           & \multicolumn{1}{c|}{0.00}           & 0.00           & 51                                                                          & \multicolumn{1}{c|}{60.55}          & \multicolumn{1}{c|}{0.19}           & \multicolumn{1}{c|}{0.25}           & 0.00           & 55                                                                           & \multicolumn{1}{c|}{45.18}          & \multicolumn{1}{c|}{0.00}           & \multicolumn{1}{c|}{0.00}           & 0.00           & 330                                                                         \\ \midrule
FGSM-PGI~\cite{jia2022prior}        & \multicolumn{1}{c|}{81.72}          & \multicolumn{1}{c|}{54.17}          & \multicolumn{1}{c|}{50.75}          & 49.0           & 73                                                                          & \multicolumn{1}{c|}{58.81}          & \multicolumn{1}{c|}{30.88}          & \multicolumn{1}{c|}{27.72}          & 25.42          & 77                                                                           & \multicolumn{1}{c|}{45.88}          & \multicolumn{1}{c|}{21.6}           & \multicolumn{1}{c|}{17.44}          & 15.50          & 458                                                                          \\ \midrule
FGSM-CKPT~\cite{kim2021understanding}       & \multicolumn{1}{c|}{\textbf{90.29}} & \multicolumn{1}{c|}{39.15}          & \multicolumn{1}{c|}{41.13}          & 37.15          & 76                                                                          & \multicolumn{1}{c|}{\textbf{60.93}} & \multicolumn{1}{c|}{15.24}          & \multicolumn{1}{c|}{16.6}           & 14.34          & 84                                                                           & \multicolumn{1}{c|}{\textbf{49.98}} & \multicolumn{1}{c|}{8.68}           & \multicolumn{1}{c|}{9.24}           & 8.10           & 495                                                                          \\ \midrule
FGSM-SDI~\cite{jia2022boosting}        & \multicolumn{1}{c|}{85.25}          & \multicolumn{1}{c|}{51.79}          & \multicolumn{1}{c|}{50.29}          & 47.91          &  83                                                                           & \multicolumn{1}{c|}{60.82}          & \multicolumn{1}{c|}{30.08}          & \multicolumn{1}{c|}{27.3}           & 25.19          & 99                                                                             & \multicolumn{1}{c|}{47.64}          & \multicolumn{1}{c|}{19.16}          & \multicolumn{1}{c|}{16.02}          & 14.10          & 565                                                                          \\ \midrule
NuAT~\cite{sriramanan2021towards} \tablefootnote{For a fair comparison, we adopt the standard ResNet18 to conduct experiments, but the original NuAT adopts ResNet18 with extra the ReLU layer at each block from the code they provide.}             & \multicolumn{1}{c|}{81.38}          & \multicolumn{1}{c|}{52.48}          & \multicolumn{1}{c|}{50.63}          & 48.70          & 104                                                                           & \multicolumn{1}{c|}{59.62}          & \multicolumn{1}{c|}{20.09}          & \multicolumn{1}{c|}{21.59}          & 11.55          & 115                                                                           & \multicolumn{1}{c|}{42.42}          & \multicolumn{1}{c|}{13.2}           & \multicolumn{1}{c|}{11.32}          & 9.56           & 660                                                                          \\ \midrule
GAT~\cite{sriramanan2021towards}            & \multicolumn{1}{c|}{80.41}          & \multicolumn{1}{c|}{51.76}          & \multicolumn{1}{c|}{49.07}          & 46.56          & 109                                                                         & \multicolumn{1}{c|}{56.07}          & \multicolumn{1}{c|}{23.0}           & \multicolumn{1}{c|}{21.93}          & 19.51          & 119                                                                           & \multicolumn{1}{c|}{41.84}          & \multicolumn{1}{c|}{13.8}           & \multicolumn{1}{c|}{11.48}          & 9.74           &  663                                                                          \\ \midrule
FGSM-GA~\cite{andriushchenko2020understanding}        & \multicolumn{1}{c|}{84.43}          & \multicolumn{1}{c|}{46.08}          & \multicolumn{1}{c|}{46.75}          & 42.63          & 178                                                                           & \multicolumn{1}{c|}{55.1}           & \multicolumn{1}{c|}{18.84}          & \multicolumn{1}{c|}{18.96}          & 16.45          & 187                                                                            & \multicolumn{1}{c|}{43.44}          & \multicolumn{1}{c|}{18.36}          & \multicolumn{1}{c|}{16.2}           & 14.28          & 1054                                                                          \\ \midrule
Free-AT (m=8)~\cite{shafahi2019adversarial}   & \multicolumn{1}{c|}{80.75}          & \multicolumn{1}{c|}{44.48}          & \multicolumn{1}{c|}{43.73}          & 41.17          & 215                                                                          & \multicolumn{1}{c|}{52.63}          & \multicolumn{1}{c|}{22.16}          & \multicolumn{1}{c|}{20.68}          & 18.57          & 229                                                                          & \multicolumn{1}{c|}{40.06}          & \multicolumn{1}{c|}{8.2}            & \multicolumn{1}{c|}{8.08}           & 7.34           & 1375                                                                          \\ \midrule
FGSM-LAW (ours) & \multicolumn{1}{c|}{81.36}          & \multicolumn{1}{c|}{\textbf{56.36}} & \multicolumn{1}{c|}{\textbf{51.4}} & \textbf{49.22} & 81                                                                           & \multicolumn{1}{c|}{58.79}          & \multicolumn{1}{c|}{\textbf{31.33}} & \multicolumn{1}{c|}{\textbf{28.47}} & \textbf{25.78} & 86                                                                           & \multicolumn{1}{c|}{47.74}          & \multicolumn{1}{c|}{\textbf{24.06}} & \multicolumn{1}{c|}{\textbf{18.65}} & \textbf{16.43} & 514                                                                          \\ \bottomrule
\end{tabular}
}
\vspace{-4mm}
\end{table*}
 
\begin{table}[t]
\centering
\caption{ Test robustness (\%) on the CIFAR-10 database using WideResNet34-10 with different adversarial training methods. Number in bold indicates the best. 
}
\label{table:cifar10_wide}
\scalebox{0.82}{

\begin{tabular}{ccccccc}
\toprule
CIFAR-10  & Clean          & PGD-10         & PGD-20         & PGD-50         & AA             & \begin{tabular}[c]{@{}c@{}}Training\\ Time (h)\end{tabular} \\ \midrule
PGD-AT~\cite{rice2020overfitting}    & 85.17          & 56.1           & 55.07          & 54.87          & 51.67          & 31.9 \\ \midrule  \midrule
FGSM-RS~\cite{wong2020fast}   & 74.3           & 42.3           & 41.2           & 40.9           & 38.4           & 5.8   \\ \midrule
FGSM-PGI~\cite{jia2022prior}  & 85.09          & 57.72          & 56.86          & 56.4           & 50.11          & 8.3 \\ \midrule
FGSM-CKPT~\cite{kim2021understanding} & \textbf{91.84} & 44.7           & 42.72          & 42.25          & 40.46          & 8.7  \\ \midrule
FGSM-SDI~\cite{jia2022boosting}  & 86.4           & 55.89          & 54.95          & 54.6           & 51.17          & 9.4 \\ \midrule
NuAT~\cite{sriramanan2021towards}      & 85.30          & 55.8           & 54.68          & 53.75          & 50.06          & 11.8 \\ \midrule
GAT~\cite{sriramanan2020guided}       & 85.17          & 56.3           & 55.23          & 54.97          & 50.01          & 12.9  \\ \midrule
FGSM-GA~\cite{andriushchenko2020understanding}   & 82.1           & 48.9           & 47.1           & 46.9           & 45.7           & 20.3 \\ \midrule
Free-AT~\cite{shafahi2019adversarial}  & 80.1           & 47.9           & 46.7           & 46.3           & 43.9           & 23.7 \\ \midrule
FGSM-LAW (Ours)      & 84.78          & \textbf{59.88} & \textbf{59.26} & \textbf{59.12} & \textbf{52.29} & 9.2 \\ \bottomrule
\end{tabular}
}
\end{table}
\begin{figure}[t]
\begin{center}
   \includegraphics[width=0.9\linewidth]{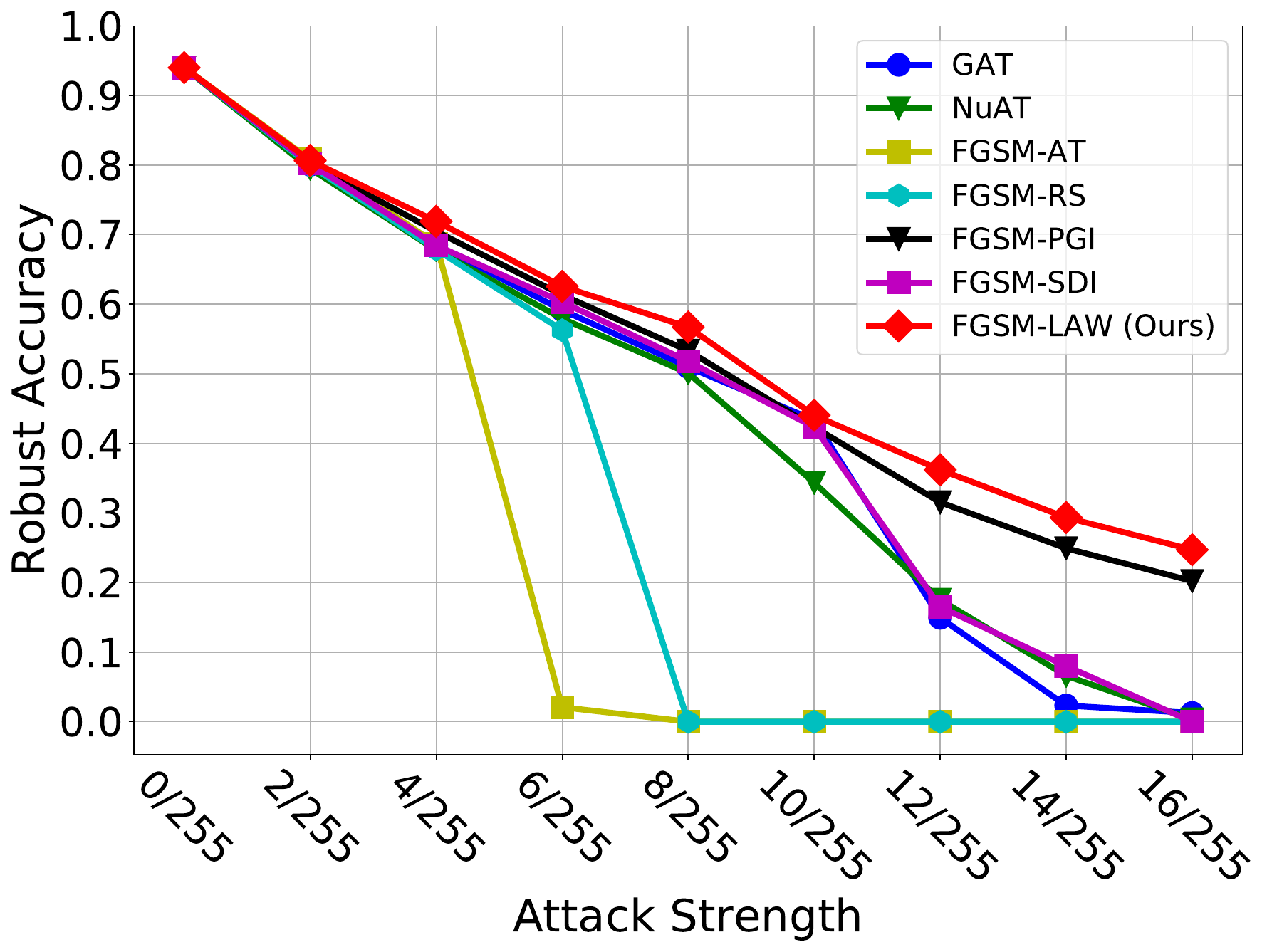}
\end{center} 
\caption{Robust accuracy VS attack strength. The adversarial robustness performance of different advanced fast adversarial training methods for CIFAR-10 with the same training setting and evaluation attack strengths under PGD-50.}
\label{fig:attack_strength}
\end{figure}
\begin{table*}[t]
\centering
\caption{ Comparisons of clean and robust accuracy (\%) and training time on the CIFAR-10 database using ResNet18 with different adversarial training methods by using a cyclic learning rate strategy. Number in bold indicates the best. 
}
\label{table:cifar10_cyclic}
\scalebox{1.0}{
\begin{tabular}{ccccccc}
\toprule
Method          & \begin{tabular}[c]{@{}c@{}}Clean\\  Best/Last\end{tabular} & \begin{tabular}[c]{@{}c@{}}PGD-10\\ Best/Last\end{tabular} & \begin{tabular}[c]{@{}c@{}}PGD-20\\ Best/Last\end{tabular} & \begin{tabular}[c]{@{}c@{}}PGD-50\\ Best/Last\end{tabular} & \begin{tabular}[c]{@{}c@{}}AA\\ Best/Last\end{tabular} & \begin{tabular}[c]{@{}c@{}}Training\\ Time (min)\end{tabular} \\ \midrule
PGD-AT~\cite{rice2020overfitting}          & 80.12/80.12                                                & 51.59/51.59                                                & 50.83/50.83                                                & 50.7/50.7                                                  & 46.83/46.83                                              & 78 \\ \midrule \midrule
FGSM-RS~\cite{wong2020fast}         & 83.75/83.75                                                & 48.05/48.05                                                & 46.47/46.47                                                & 46.11/46.11                                                & 42.92/42.92                                              & 15  \\ \midrule
FGSM-PGI~\cite{jia2022prior}       & 80.68/80.68                                                & 52.48/53.48                                                & 51.69/51.69                                                & 51.5/51.5                                                  & 46.65/46.65                                              & 22 \\ \midrule
FGSM-CKPT~\cite{kim2021understanding}       & \textbf{89.08/89.08}                                       & 40.47/40.47                                                & 38.2/38.2                                                  & 37.69/37.69                                                & 35.66/35.66                                              & 23 \\ \midrule
FGSM-SDI~\cite{jia2022boosting}        & 82.08/82.08                                                & 51.63/51.63                                                & 50.65/50.65                                                & 50.33/50.33                                                & 46.21/46.21                                              & 24 \\ \midrule
NuAT~\cite{sriramanan2021towards}           & 76.23/76.23                                                & 51.52/51.52                                                & 50.81/50.81                                                & 50.64/50.64                                                & 46.33/46.33                                              & 30 \\ \midrule
GAT~\cite{sriramanan2020guided}             & 81.91/81.91                                                & 50.43/50.43                                                & 49.82/49.82                                                & 49.62/49.62                                                & 45.24/45.24                                              & 33 \\ \midrule
FGSM-GA~\cite{andriushchenko2020understanding}         & 80.83/80.83                                                & 48.76/48.76                                                & 47.83/47.83                                                & 47.54/47.54                                                & 43.06/43.06                                              & 53  \\ \midrule
Free-AT (m=8)~\cite{shafahi2019adversarial}   & 75.22/75.22                                                & 44.67/44.67                                                & 43.97/43.97                                                & 43.72/43.72                                                & 40.30/40.30                                              & 58 \\ \midrule
FGSM-LAW (ours) & 80.44/80.44                                                & \textbf{53.54/53.54}                                       & \textbf{52.91/52.91}                                       & \textbf{52.73/52.73}                                       & \textbf{46.93/46.93}                                     & 24 \\ \bottomrule
\end{tabular}
}
\end{table*}

\begin{table}[t]
\centering
\caption{ Test robustness (\%) on the  ImageNet database using ResNet50 with different adversarial training methods. Number in bold indicates the best. 
}
\label{table:ImageNet}
\scalebox{1}{
\begin{tabular}{cccc}
\toprule
Method   & Clean          & AA             & Time (hour) \\ \midrule
PGD-AT \cite{madry2017towards}   & 64.02          & 34.96 &  211.2\\ \midrule \midrule
Free-AT \cite{shafahi2019adversarial}  & 59.96          & 28.58          & 127.7 \\ \midrule
FGSM-RS \cite{wong2020fast}  & 55.62          & 26.24          & 44.5   \\ \midrule
FGSM-LAW (Ours) & \textbf{66.14} & \textbf{30.12} & 62.3 \\ \bottomrule
\end{tabular}
}
\vspace{-4mm}
\end{table}
\begin{table*}[t]
\centering
\caption{ Comparisons of clean and robust accuracy (\%) and training time on the CIFAR-10 database using ResNet18. Number in bold indicates the best. 
 }
\label{table:NuAT_WA}
\scalebox{1}{
\begin{tabular}{ccccccccc}
\toprule
                           & Method   & Clean          & PGD-10         & PGD-20        & PGD-50         & C\&W           & AA             & \begin{tabular}[c]{@{}c@{}}Training \\ Time(min)\end{tabular} \\ \midrule
\multirow{2}{*}{CIFAR-10}  & NuAT-WA~\cite{sriramanan2021towards}   & \textbf{82.21} & 54.76          & 54.07         & 53.87          & 52.07          & 50.54          & 95                                                            \\ \cmidrule(l){2-9} 
                           & FGSM-LAW (ours) & 81.8           & \textbf{58.13} & \textbf{57.6} & \textbf{57.47} & \textbf{52.46} & \textbf{50.64} & 81                                                            \\ \midrule
\multirow{2}{*}{CIFAR-100} & NuAT-WA~\cite{sriramanan2021towards}   & \textbf{61.6}  & 27.57          & 24.04         & 21.77          & 22.58          & 15.11          & 105                                                           \\ \cmidrule(l){2-9} 
                           & FGSM-LAW (ours) & 59.2           & \textbf{31.95} & \textbf{31.4} & \textbf{31.3}  & \textbf{28.27} & \textbf{25.72} & 86                                                            \\ \bottomrule
\end{tabular}
}
\end{table*}

\begin{table*}[t]
\centering
\caption{Comparisons of clean and robust accuracy (\%) and training time on the CIFAR-10 of the best and last checkpoints using different adversarial training with different model weight averaging methods under $\ell_{\infty}= 8/225$. Number in bold indicates the best. . 
}
\label{table:FGSM_PGI}
\scalebox{1}{
\begin{tabular}{cccccccc}
\toprule
Method           & \begin{tabular}[c]{@{}c@{}}Clean\\ Best/Last\end{tabular} & \begin{tabular}[c]{@{}c@{}}PGD-10\\ Best/Last\end{tabular} & \begin{tabular}[c]{@{}c@{}}PGD-20\\ Best/Last\end{tabular} & \begin{tabular}[c]{@{}c@{}}PGD-50\\  Best/Last\end{tabular} & \begin{tabular}[c]{@{}c@{}}C\&W\\ Best/Last\end{tabular} & \begin{tabular}[c]{@{}c@{}}AA\\  Best/Last\end{tabular} & {\begin{tabular}[c]{@{}c@{}}Training \\ Time (min)\end{tabular}} \\ \midrule
FGSM-PGI~\cite{jia2022prior}         & 81.72/81.72                                               & 55.18/55.18                                                & 54.36/54.36                                                & 54.17/54.17                                                 & 50.75/50.75                                              & 49.0/49.0                                               & 73  \\ \midrule \midrule

FGSM-PGI-EMA ~\cite{jia2022prior}    & 80.6/82.3                                              & 55.73/55.18                                                & 55.02/54.48                                                &   55.0/54.14                                        &            50.86/50.74                                        & 49.23/49.08                                              & 74 \\ \midrule
FGSM-PGI-Auto-EMA (ours) & 80.71/81.0                                               & \textbf{55.87/55.85}                                       & \textbf{55.32/55.31}                                       & \textbf{55.11/55.15}                                        & \textbf{51.07/51.0}                                     & \textbf{49.66/49.56}                                    & 74 \\ \bottomrule
\end{tabular}
}
\end{table*}

\subsection{Detailed hyper-parameter settings}
\label{appendix:hyper-parameter}
There are two hyper-parameters in the proposed FGSM-LAW, \emph{i.e,}
the hyper-parameter of the Lipschitz regularization $\lambda$ and the hyper-parameter of the auto weight averaging $\Delta$. Using ResNet18 on CIFAR-10, we adopt the FGSM-BR combined with Lipschitz regularization and auto weight averaging to determine the optimal hyper-parameters, respectively. As for the hyper-parameter $\lambda$, the results are shown in Table~\ref{table:HP_regular}. It can be observed that when $\lambda=12$, the proposed Lipschitz regularization can achieve the best model robustness  under all attack scenarios. In detail, under AA attack, the proposed Lipschitz regularization achieves an accuracy of about 48\%. Hence, the hyper-parameter $\lambda$ of the proposed Lipschitz regularization is set to 12. As for the hyper-parameter $\Delta$, the results are shown in Table~\ref{table:HP_WA}. It can be observed that when $\Delta =0.82$, the proposed auto weight averaging can achieve the best model robustness  under C\&W and AA attack scenarios. Considering the clean accuracy on clean samples, the hyper-parameter $\Delta$ of the proposed auto weight averaging is set to 0.82.



\subsection{Comparisons With State-of-the-Art Methods}
To evaluate the proposed method, we compare the proposed method with several state-of-the-art
fast adversarial training methods \emph{i.e,} Free-AT\cite{shafahi2019adversarial}, FGSM-RS~\cite{wong2020fast}, FGSM-GA~\cite{andriushchenko2020understanding}, GAT~\cite{sriramanan2020guided}, FGSM-CKPT~\cite{kim2021understanding}, NuAT~\cite{sriramanan2021towards}, FGSM-SDI~\cite{jia2022boosting}, FGSM-PGI~\cite{jia2022prior} and an advanced SAT method (\emph{i.e,} PGD-AT~\cite{rice2020overfitting}). Note that we adopt the training settings reported in their original works to train these AT models. 

\subsubsection{Results on CIFAR-10}
As for CIFAR-10, we use ResNet18 as the backbone. The results are shown in Table~\ref{table:experiments}.  It can be observed that compared with the advanced PGD-AT, the proposed method achieves better model robustness under all attack scenarios and requires much less training time. 
Compared with other advanced fast adversarial training methods, as for the best and last checkpoints, the proposed method achieves the best model robustness under all attack scenarios. Specifically, under the PGD-50 attack, the previous most effective fast adversarial training methods achieve an accuracy of about 54\%, while the proposed method achieves an accuracy of about 56\%. 
In terms of the training cost, the proposed method requires the minimum training time except for FGSM-RS which meets Catastrophic Overfitting. The training time of the proposed method is the same as FGSM-SDI, which is about 1.3 times faster than GAT and NuAT, 2.2 times faster than FGSM-GA and 2.7 times faster than Free-AT. FGSM-PGI requires huge memory costs to store the historical adversarial perturbations. But our method requires the same memory cost as the other fast adversarial training methods.
Different from previous fast adversarial training variants, the proposed method not only prevents Catastrophic Overfitting but also improves the model robustness without extra training time and memory cost. Moreover, following the default settings, we compare the proposed method with several advanced fast adversarial training methods under larger $\ell_{\infty}$ distance metric. As shown in Fig.~\ref{fig:attack_strength}, previous fast adversarial training methods could lead to Catastrophic Overfitting but the proposed method can prevent Catastrophic Overfitting and achieve significant adversarial robustness improvement.

\par Besides, we also adopt a larger architecture (WideResNet34-10) as the backbone to conduct comparative experiments on CIFAR-10. 
The results are shown in Table~\ref{table:cifar10_wide}. It can be observed that compared with PGD-AT, the proposed FGSM-LAW achieves better model robustness. Particularly, under AA attack, PGD-AT achieves an
accuracy of about 51\%, while the proposed FGSM-LAW achieves an accuracy of about 52\%. Compared with previous fast adversarial training methods, the proposed method achieves the best robustness performance under all attack scenarios. In terms of the training cost, 
we observe similar phenomenons as ResNet18 used as the backbone.

\par Moreover, the comparative experiments are also implemented by using a cyclic learning rate strategy on CIFAR-10. Following the training settings (\cite{wong2020fast, jia2022prior}, the maximum learning rate of FGSM-GA~\cite{andriushchenko2020understanding} and FGSM-CKPT~\cite{kim2021understanding} is set to 0.3. The maximum learning rate of other fast adversarial training methods is set to 0.2. The results are shown in Table~\ref{table:cifar10_cyclic}. 
It can be observed that compared with PGD-AT, the proposed FGSM-LAW  achieves better model robustness performance under all adversarial attack scenarios and require less training cost. And compared with previous fast adversarial training methods, the proposed FGSM-LAW can achieve the best adversarial robustness under
all adversarial attack scenarios on the best and last checkpoint. In terms of the training cost, 
we observe similar phenomenons as the models trained with the multi-step learning rate strategy.

\subsubsection{Results on CIFAR-100}  
As for CIFAR-100, ResNet18 is used as the backbone network. The result is shown in Table~\ref{table:experiments}. Compared with PGD-AT, the proposed method achieves better model robustness under most attack scenarios and costs less training time. Even under AA attack, PGD-AT obtains an accuracy of about 25.48\%, but our method obtains an accuracy of about 25.78\%.
Our method achieves the best model robustness under all attack scenarios among the fast adversarial training methods. In terms of the training cost, similar results are observed on CIFAR-10. 

\subsubsection{Results on Tiny ImageNet and ImageNet}  
As for Tiny ImageNet, following the previous works~\cite{andriushchenko2020understanding,jia2022boosting,jia2022prior} PreActResNet18 is used as the backbone network. The result is shown in Table~\ref{table:experiments}. Compared with PGD-AT, the proposed method achieves better adversarial robustness under all attack scenarios with less training time. In detail, under AA attack, PGD-AT achieves an accuracy of about 12.84\%, but the proposed method achieves an accuracy of about 16.43\%.
And compared with previous fast adversarial training methods, the proposed method achieves the best model robustness under all attack scenarios. In terms of the training cost, similar results are observed on CIFAR-10 and CIFAR-100. Following the training settings ~\cite{shafahi2019adversarial,wong2020fast}, we adopt ResNet50 as the backbone to conduct comparative experiments. The maximum perturbation strength $\epsilon$ is set to 4/255. We compare the proposed FGSM-LAW with PGD-AT \cite{madry2017towards}, Free-AT \cite{shafahi2019adversarial} and FGSM-RS \cite{wong2020fast}. 
The results are shown in Table~\ref{table:ImageNet}. Compared with PGD-AT, our FGSM-LAW achieves higher clean accuracy and comparable robustness under AA attack. Compared with Free-AT and FGSM-RS, the proposed FGSM-LAW achieves the best clean and robust accuracy.

\subsubsection{Comparisons with state-of-the-art robustness model} 

\noindent  \textbf{Comparisons With NuAT-WA.}
NuAT-WA uses CyclicLR to improve robustness, achieving state-of-the-art results in adversarial robustness. To conduct a fair comparison, we adopt the same training settings to conduct FGSM-LAW. The results are shown in Table~\ref{table:NuAT_WA}. Compared to NuAT-WA, our proposed method demonstrates better adversarial robustness under all attack scenarios on both CIFAR-10 and CIFAR-100 datasets. Additionally, the proposed method incurs a lower training cost than NuAT-WA.

\noindent  \textbf{Comparisons With FGSM-PGI-WA.}
The proposed auto weight averaging method can be combined with FGSM-PGI to improve the adversarial robustness of the original FGSM-PGI. We adopt FGSM-PGI with the original weight averaging as the baseline. The results are shown in Table~\ref{table:FGSM_PGI}. It can be observed that FGDM-PGI combined with the proposed Auto-EMA can achieve the best adversarial robustness under all attack scenarios. In detail, under AA attack, FGSM-PGI-EMA achieves an
accuracy of about 49.08\%, while the proposed FGSM-PGI-Auto-EMA achieves an accuracy of about 49.56\% with the same training time, which validates the effectiveness of the proposed method.

\begin{table}[!htb]
\centering
\caption{ Ablation study on CIFAR-10. Clean and robust accuracy (\%) and training time on the CIFAR-10 database using ResNet18 under $\ell_{\infty}= 8/225$. Number in bold indicates the best. 
}
\label{table:ablation}
\scalebox{0.72}{
\begin{tabular}{cccccccc}
\toprule
L-Regular (Ours) & Cutout & A-WA (Ours) & Clean          & PGD-50         & C\&W             & AA             & \begin{tabular}[c]{@{}c@{}}Training\\ Time (min)\end{tabular} \\ \midrule
                   &        &                      & \textbf{86.82} & 44.23          & 45.89         & 42.37           &     51   \\ \midrule  \midrule 
     \checkmark    &        &                      & 82.97          & 54.29          & 50.41          & 48.05          &    73    \\ \midrule
    \checkmark     &   \checkmark     &                      & 82.55          & 55.08          & 50.69          & 48.63          &     80  \\ \midrule
    \checkmark     &        &       \checkmark               & 82.56          & 55.64          & 51.22          & 48.80          &    74  \\ \midrule
     \checkmark    &   \checkmark     &         \checkmark             & 81.34          & \textbf{56.36} & \textbf{51.4} & \textbf{49.22} &   81   \\ \bottomrule
\end{tabular}
}
\end{table}
\subsection{Ablation Study}
\label{sec:ablation}

In the proposed method, we propose a Lipschitz regularization (L-regular) term and an auto weight averaging (A-WA) with Cutout augmentation to conduct fast adversarial training. To validate the
effectiveness of each element, we conduct an ablation study on CIFAR-10 by using ResNet18. 
The result is shown in Table~\ref{table:ablation}. We use the clean accuracy on clean images and the robust accuracy on different adversarial attack methods as the evaluation metric on the last checkpoint. 
Analyses are summarized as follows. First, only using the proposed regularization, the performance of the model robustness under all attack scenarios significantly improves. 
Incorporating our regularization with cutout augmentation, the performance of the model robustness further improves. Incorporating our regularization with the proposed weight averaging, the proposed method can better model robustness. Second, using all terms can obtain the best robust performance, which indicates that the three elements are compatible, and combining them can achieve the best performance. 
\section{Conclusion}
This work conducts a thorough analysis of existing works in the field of fast adversarial training in terms of adversarial robustness and training cost. We revisit the effectiveness and efficiency of the fast adversarial training techniques in preventing catastrophic overfitting. To further improve the performance of fast adversarial training, we propose a novel regularization method motivated by the Lipschitz constraint. Moreover, we explore the effect of data augmentation and weight averaging in fast adversarial training and propose an effective auto weight averaging method for fast adversarial training to improve robustness. 
By assembling these techniques, we conclude our fast adversarial
training method equipped with the proposed Lipschitz regularization and
our auto weight averaging, called FGSM-LAW. Extensive experimental evaluations demonstrate that the proposed method outperforms state-of-the-art fast adversarial training methods with less training costs. 



\bibliographystyle{IEEEtran}
%

\bibliography{egbib}


 




\vfill

\end{document}